\documentclass{article}

\usepackage[preprint]{nips_2018}

\usepackage[utf8]{inputenc} 
\usepackage[T1]{fontenc}    
\usepackage{hyperref}       
\usepackage{url}            
\usepackage{booktabs}       
\usepackage{nicefrac}       
\usepackage{microtype}      
\usepackage{amsfonts}
\usepackage{amsmath}
\usepackage{amssymb}
\usepackage{wrapfig}
\usepackage{mathtools}
\usepackage{subcaption}
\usepackage{verbatim}
\usepackage{hyperref}
\usepackage[capitalize,nameinlink]{cleveref}
\usepackage[inline]{enumitem}   
\usepackage{makecell, tabularx}
\usepackage{floatpag}

\newif\ifanony  
\anonyfalse

\usepackage{natbib}
\bibpunct{(}{)}{;}{a}{,}{,}

\AtBeginDocument{

}

\DeclarePairedDelimiter{\pdelims}{(}{)}
\DeclarePairedDelimiter{\bracedelims}{\{}{\}}
\newcommand{\of}[1]{\pdelims*{#1}}
\newcommand{\bof}[1]{\bracedelims*{#1}}

\newcommand{\amplify}[1]{\operatorname{Amplify^H}\of{#1}}
\newcommand{\amplifyp}[1]{\operatorname{Amplify^{H^{\prime}}}\of{#1}}

\DeclareMathOperator{\layernorm}{LayerNorm}
\DeclareMathOperator{\batchnorm}{BatchNorm}
\DeclareMathOperator{\mlp}{MLP}
\DeclareMathOperator{\attention}{Attention}

\title{Supervising strong learners\\by amplifying weak experts}
\author{
  Paul Christiano \\
  OpenAI \\
  \texttt{paul@openai.com} \\
  \And
  Buck Shlegeris \thanks{Work done while at OpenAI.}\\
  \texttt{bshlegeris@gmail.com} \\
  \And
  Dario Amodei \\
  OpenAI \\
  \texttt{damodei@openai.com} \\
}
\date{\today}

\begin{document}

\maketitle

\begin{abstract}%
Many real world learning tasks involve complex or hard-to-specify objectives, and using an easier-to-specify proxy can lead to poor performance or misaligned behavior.  One solution is to have humans provide a training signal by demonstrating or judging performance, but this approach fails if the task is too complicated for a human to directly evaluate.  We propose Iterated Amplification, an alternative training strategy which progressively builds up a training signal for difficult problems by combining solutions to easier subproblems. Iterated Amplification is closely related to Expert Iteration \citep{exit, agz}, except that it uses no external reward function. We present results in algorithmic environments, showing that Iterated Amplification can efficiently learn complex behaviors.
\end{abstract}

\section{Introduction}

If we want to train an ML system to perform a task, we need to be able to evaluate how well it is doing. Whether our training signal takes the form of labels, rewards, or something else entirely, we need some way to generate that signal.

If our goal can be evaluated automatically,
such as winning a game of Go,
or if we have an algorithm that can generate examples of correct behavior, then generating a training signal is trivial.
In these cases we might say that there is an ``algorithmic'' training signal.

Unfortunately, most useful tasks don’t have an algorithmic training signal.
So in current applications of machine learning, humans often provide the training signal.
This can be done by having a human demonstrate the task, for example  labeling an image or teleoperating a robot,
or by learning a reward function from human judgments.
For these classes of tasks, we could say there is a “human” training signal.

However, there are harder tasks for which we can’t compute demonstrations or rewards even with human assistance,
and for which we currently have no clear method to get a meaningful training signal.
Consider making economic policy decisions,
advancing the scientific frontier,
or managing the security of a large network of computers.
Some of these tasks are ``beyond human scale'' -- a single human
can’t perform them and can’t make sense of their massive observation space well enough to judge the behavior of an agent.
It may be possible for a human to judge performance in the very long run
(for example, by looking at economic growth over several years),
but such long-term feedback is very slow to learn from.
We currently have no way to learn how to perform such tasks much better than a human.

The overall situation is depicted in Table 1, which shows six different combinations of training signal source and problem formulation (supervised learning or RL).
The bulk of ML practice operates in the top center box (supervised learning from human labels), the bottom left box (RL with a scripted reward), and sometimes the top left box (supervised learning of algorithms).  The bottom center box (RL from a human training signal) is beginning to be explored, and includes inverse reinforcement learning \citep{Ng00, irl2, gcl} and RL from human feedback \citep{tamer, prosthetic, coach, humanfeedback}.  At present there seems to be no general method to handle problems in the bottom right or top right.

It seems desirable to expand the range of tasks for which we can get a training signal, for two reasons.  First, it would enable ML systems to perform new tasks.  SL and RL are very powerful methods when we can get a training signal, so making them applicable to tasks that humans can't directly judge or perform could have a big impact.  Second, better specification of complex goals and targets may be vital to building robustly beneficial AI systems.  In practice, when an accurate training signal would be ``beyond human scale,'' we often instead find a short-term proxy that is correlated with what we want. But aggressively optimizing that proxy can lead to pathological behavior \mbox{\citep{creativity, faulty, Amodei16}},
an example of Goodhart’s Law.\footnote{``When a measure becomes a target, it ceases to be a good measure''}
For example, we might find that user-reported satisfaction (which we can easily measure) is a good proxy for long-term benefit to society (which is very complicated),
but if we maximize it with RL our agent may maintain fraudulent appearances
or effectively manipulate users into providing high ratings.
At large scales this kind of pathology could lead to systemic crashes,
and a mismatch between proxies and our real preferences is
a major source of concerns about the safety of future powerful AI systems \mbox{\citep{superintelligence}}.

In this paper we propose a general framework for building up a training signal on complex tasks by decomposing them (with AI assistance) into simpler tasks for which we have a human or algorithmic training signal.  In our experiments we apply the framework with a number of simplifications (see \cref{sec:future}) to relatively simple tasks, as a first step towards addressing the problems described above.

\begin{table}
\centering
\caption{Example problems which require different kinds of training signal.}\label{feedbacktypes}
\begin{tabular}{cccc}
\toprule
Training signal: & Algorithmic & Human & Beyond human \\
\midrule
\midrule
\makecell{Supervised \\learning} &
\makecell{Learning data structures \\
\includegraphics[width=0.2\textwidth]{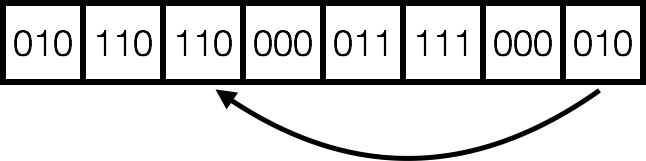}
}&
\makecell{Image classification \\
\includegraphics[width=0.15 \textwidth]{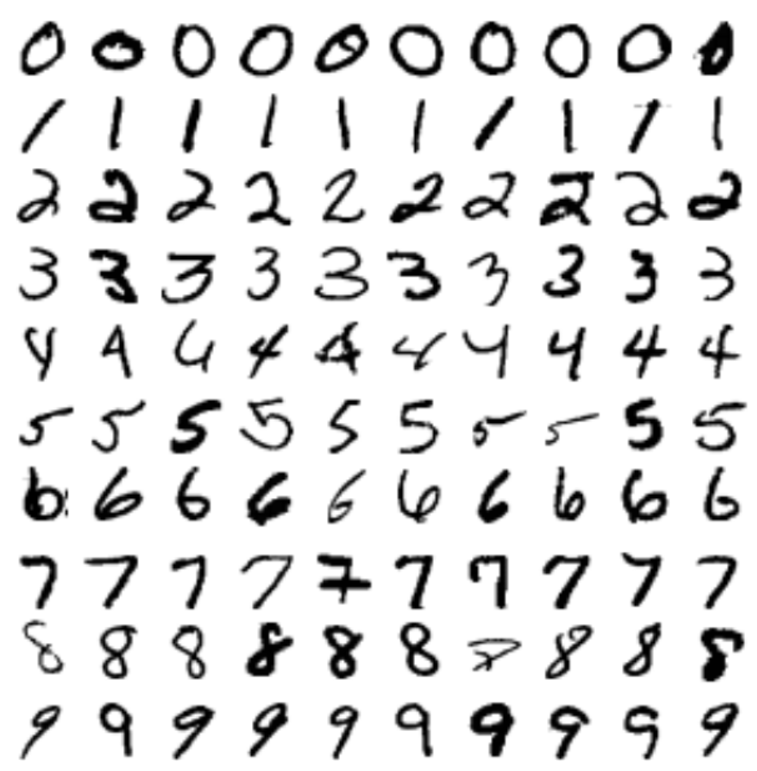}
} &
\makecell{Long-term prediction \\
\includegraphics[width=0.2\textwidth]{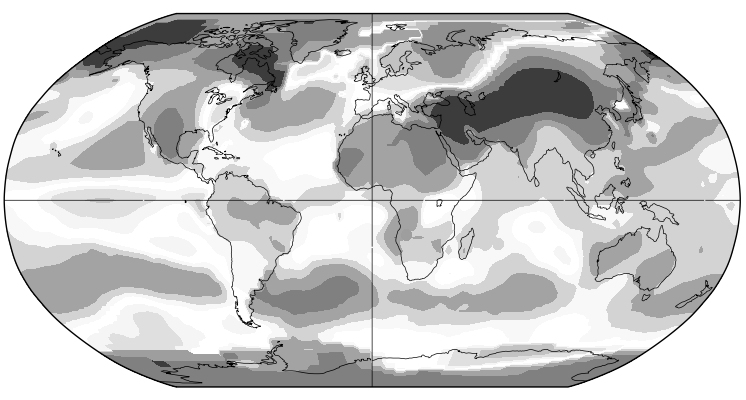}
}
\\
\midrule
\makecell{Reinforcement\\learning} &
\makecell{Playing games\\
\includegraphics[width=0.2\textwidth]{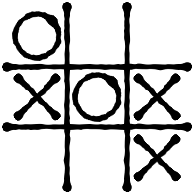}
}
 & \makecell{Driving ``well''\\
\includegraphics[width=0.15 \textwidth]{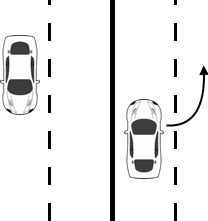}
} &
\makecell{Designing transit system \\
\includegraphics[width=0.2 \textwidth]{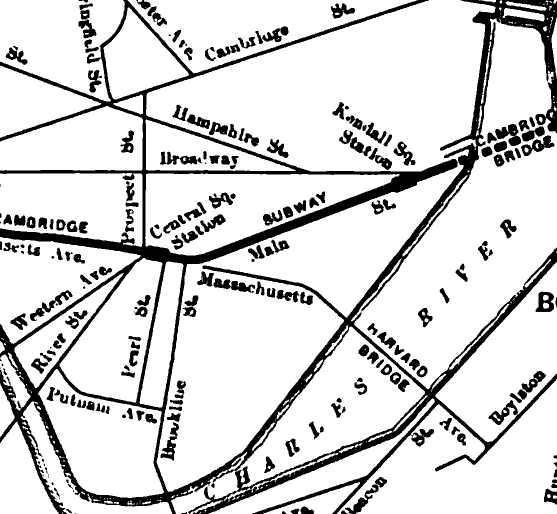}} \\
\bottomrule
\end{tabular}
\end{table}


\subsection{Our method: Iterated Amplification}
\label{sec:methodintro}

We propose a new method, Iterated Amplification,
for a human expert $H$ to train an ML agent $X$.
Rather than having $H$ demonstrate or evaluate the target behavior on their own,
we allow them to invoke several copies
of the current agent $X$ to help them.
We write $\amplify{X}$ for the composite system, consisting of $H$ and several copies of $X$ working together to solve a problem.
The agent $X$ then learns from $\amplify{X}$
in the same way that it would traditionally
learn from $H$ alone.

To instantiate this framework we make three design decisions:
\begin{itemize}
\item What set of tasks do we train $X$ to solve?
In order for $X$ to be a useful assistant,
we need to choose a sufficiently broad set of tasks.
In this article,
we will focus on question-answering.
\item
How do we construct $\amplify{X}$?
In this article, we focus on delegation:
$\amplify{X}$ answers a question $Q$ by having
$H$ identify a sequence of useful subquestions,
using $X$ to compute a subanswer to each subquestion,
and having $H$ decide how to answer $Q$
after seeing the subanswers.
\item
How does $X$ learn from $\amplify{X}$?
In this article, we focus on supervised learning: $X$ is an autoregressive
model trained to predict $\amplify{X}$'s output.
Future work could instead use imitation learning,
or use $\amplify{X}$
to define a reward function
that $X$ maximizes with RL.
\end{itemize}

Initially $X$ behaves randomly, so
$\amplify{X}$ is essentially equivalent
to $H$ and we are effectively
learning from an expert.
Over time the agent $X$ becomes more powerful and the role of the expert
transitions into ``coordinating'' several copies of $X$ to solve the problem better
than a single copy could solve it.
(Once $X$ is very sophisticated,
even tasks like
``identify a useful subquestion'' might be delegated.)
As long as it is possible for multiple
agents to collaboratively solve
problems more effectively than a single
agent (perhaps using human expertise
to coordinate their efforts),
then $\amplify{X}$ can outperform
$X$ and hence provide a useful training signal. We discuss this assumption in \cref{sec:decomposition}.

The human must be involved in this process
because there is no external objective to guide learning---the
objective is implicit in the way that the human coordinates the copies of $X$.
For example,
we have no external measure of
what constitutes a ``good''
answer to a question,
this notion is only implicit in how a human decides to combine the answers
to subquestions (which usually involves
both facts and value judgments).
Our goal is for $X$ to learn the goal
at the same time that it learns
to behave competently.
This is in contrast
with the alternative approach of specifying a reward function
and then training a capable
agent to maximize that reward function.

\subsection{Outline}

In \cref{sec:method} we describe
Iterated Amplification and our implementation in more detail.
In \cref{sec:related} we compare
our approach to prior work.
In \cref{sec:experiments}
we describe our experimental results,
showing that 
Iterated Amplification can be stable and efficient despite
the non-stationary training signal and lack of external objective.
In \cref{sec:decomposition} we explain
why we believe that decomposability
is a realistic assumption for complex tasks in the real world.

\section{Detailed instantiation of Iterated Amplification}
\label{sec:method}

\subsection{Predicting the human behavior}

In order to reduce the burden
on the human expert $H$,
we train a ``human predictor'' $H'$,
and use this predictor to generate
training data rather than consulting $H$ directly.
That is,
we train $H'$
to imitate the role of $H$
when computing $\amplify{X}$,
and we train $X$
using $\amplifyp{X}$
rather than using $\amplify{X}$
directly.

Because $H'$ is only learning
how to identify subquestions
and combine subanswers,
rather than solving an entire task,
we expect to train it with much less data.

Note that $H'$ needs to predict
how $H$ will respond to subanswers
provided by $X$.
Because $X$ is changing,
this distribution is non-stationary,
and so we need to continuously update
$H'$ throughout the training process.

\subsection{Training overview}

\newcommand{\dist}{\mathcal{D}}

We train an agent $X$ to answer questions
from some distribution $\dist$.

\begin{figure}
\centering
\includegraphics[width=1.1\textwidth]{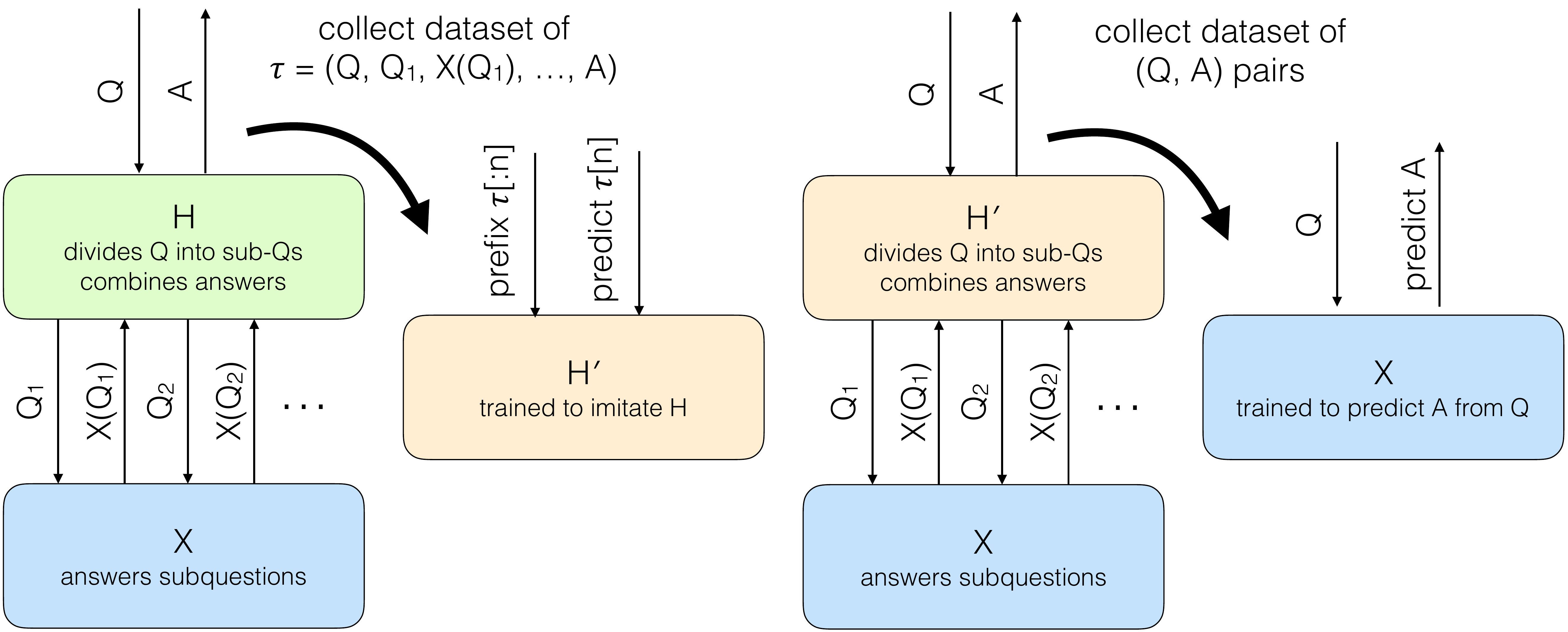}
\caption{Schematic of our Iterated Amplification implementation.}\label{fig:process}
\end{figure}

Our training process, depicted in \cref{fig:process}, involves running four processes in parallel:

\begin{enumerate}
\item
We repeatedly sample a question $Q \sim \dist$,
use $\amplify{X}$ to answer that question,
and record every decision made by $H$ during the process. That is,
$H$ finds a subquestion $Q_1$
that would help them answer $Q$,
and we compute the answer $A_1 = X(Q_1)$.
We repeat this process $k$ times,
where $k$ is a fixed parameter,
and then $H$ computes an answer
$A$. We store the transcript
$\tau = \of{Q, Q_1, A_1, \ldots, Q_k, A_k, A}$.
\item We train a model $H'$ to predict the decisions made by $H$
in each of these transcripts,
i.e. to predict subquestions $Q_i$ and final answers $A$.
\item We repeatedly sample a question $Q \sim \dist$,
use $\amplifyp{X}$ to answer that question,
and record the resulting $(Q, A)$ pairs.
\item $X$ is trained by supervised learning on these $(Q, A)$ pairs.
\end{enumerate}

\subsection{Dynamics of training}

The behavior of the agent $X$ develops over the course of training:

\begin{itemize}
\item Initially $X$ answers questions randomly.
When the human asks subquestions they frequently receive incoherent
or useless subanswers,
\item The human is able to answer some questions without any help from $X$,
and eventually $X$ learns to copy these simple answers.
\item Once $X$ is able to provide simple answers,
the human is able to provide slightly better answers
by breaking them into simple pieces.
Then $X$ learns to provide slightly better answers.
\item This process continues,
with $X$ gradually expanding the set of queries it can answer
and gradually improving the answers it provides.
At each point in training, $\amplify{X}$ is modestly smarter than X working on its own,
and $X$ chases this moving target.
\end{itemize}

If all goes well, at the end of this process we’re left with an agent that
``approximates'' the behavior of an exponentially large team of copies of $H$.
The hierarchical decomposition itself is discarded as an artifact of training,
and the actual procedure learned by the agent
will generally not mirror the structure used in training.

\subsection{Questions with context}
\label{sec:context}

In practice, questions often take the form of a
very large \emph{context} (e.g. a hundred-page design document)
and a relatively small context-conditional question
(e.g. ``what are the largest risks when implementing this design?'').
In particular, this is the case in the experiments
reported in \cref{sec:experiments}.

Answering a question requires understanding the entire context,
but it may be possible to \emph{decompose} a question
without looking at the entire context.
This allows us to apply Iterated Amplification
to tasks where the context is too large for a human expert
to observe directly.
Instead, we can give $H$ the ability
to look at small parts of the context as needed.
Hopefully, $H$ can decompose a question
into pieces that depend on smaller and smaller parts of the context,
until arriving at questions that depend on only isolated facts
from the context.

Large contexts also facilitate
an important trick for accelerating training.
In many settings, almost all of the work
of answering a question is actually about
understanding the context,
and it is possible to ask many different
questions about a single context.

We divide $X$ into two phases,
a context-encoding phase
and a question-answering phase.
During training,
rather than sampling a single question
we sample a context together with multiple questions about that context.
We reuse the work of the context-encoding
phase across all of these questions.
Similarly, when computing $\amplify{X}$,
we reuse the context-encoding
work between all of the subquestions
that $X$ needs to answer.
In our experiments,
this speeds up training by an order of magnitude.

\subsection{Model architecture}

We implement $X$
as an encoder-decoder architecture with self-attention,
closely following the Transformer architecture \citep{transformer}:
\begin{itemize}
\item We represent the context as a set of facts,
each of which is a sequence of tokens.
\item We embed each token using a look-up table.
We embed facts by concatenating the token embeddings
and applying a linear projection.
\item We apply the Transformer encoder to the embedded facts.
Our only change to the architecture from \citep{transformer}
is inserting batchnorm after each MLP.
\item We embed questions in the same way we embed facts,
then apply the Transformer decoder
to a batch of questions
(omitting the self-attention altogether because it would correspond to interactions amongst questions, which ought to be independent).
\item An autoregressive MLP generates a sequence of symbols conditioned on the result of the Transformer decoder. It generates symbols either by outputting
a set of logits or by choosing to copy a symbol from the context
(as in pointer networks \citep{pointer}).
\end{itemize}

The human-predictor $H'$ is also a Transformer decoder
augmented with the ability to copy symbols from previous steps.
$H'$ operates on sequences of questions and answers---like $H$,
it never observes the entire context.

Details of our model architecture are described in \cref{app:details}.

\section{Related Work}
\label{sec:related}

\textbf{Expert Iteration:}
our method is very similar to Expert Iteration (ExIt) \citep{exit}
and AlphaZero \citep{agz, az}
and has recently achieved strong performance in the board games Hex, Go, Chess, and Shogi.
ExIt is itself closely analogous to the Bellman update in Q learning,
and all of these can be viewed as analogs of dynamic programming where neural networks replace lookup tables.

The key difference between our work and ExIt is the lack of an external objective.
In ExIt, the expert is produced by a search algorithm that optimizes
an external objective.
Our contribution is to show that that a similar training process can be used even when the
task definition is only implicit in the decomposition and recomposition strategy.

\textbf{Inverse reinforcement learning:} by observing
human behavior and inferring the underlying reward function that they are optimizing,\citep{irl, cirl}
inverse reinforcement learning could also potentially learn reward functions for tasks that are
too challenging for humans.
Handling such tasks requires a sufficiently accurate model of human cognition
to predict what humans ``would'' prefer if we relaxed their cognitive limitations;
in addition to being extremely complex,
such a model is not identifiable,
because we never observe the ground truth about human preferences.
Iterated Amplification is an alternative strategy
that does not require solving this challenging model specification problem.

\textbf{Debate:} training AI systems to debate each other
\citep{debate}
is another possible approach to training question-answering
systems where a human expert cannot evaluate answers directly.
Both debate and Iterated Amplification involve a recursive
structure where AI systems help humans address relevant subquestions.
The largest conceptual difference is that in Iterated Amplification
each subquestion is answered by an independent copy of $X$
trained by $\amplify{X}$,
while in a debate the subquestions are answered by one of the debaters
(who are trained to defend a particular answer
to the top-level question).

\textbf{Algorithm learning:}
our problem differs from traditional work on learning algorithms
\citep{dnc, neuralgpu, neuralprogrammer}
because we \textbf{don't} assume that we have access to ground truth labels.

\textbf{Recursive model architectures:}
our work differs from recursive model architectures \citep{recursion, divideandconquer},
in that the learned model doesn't have a recursive structure.
The recursive decomposition is used only to generate training data,
and even then only a single step of decomposition is performed in each iteration.

So the trained agent might end up solving the task in a totally different way
from the decomposition used by the human,
and in particular it may learn heuristics that treat the problem holistically.

This flexibility is important to the applicability of our method.
It is often possible to divide a task into easier parts,
but dividing interesting problems into pieces and solving them independently
can be much less efficient than considering the problem holistically.

\section{Experiments}
\label{sec:experiments}

\subsection{Tasks}

We study Iterated Amplification in a set of 5 toy algorithmic tasks.
For each task, the agent $X$ is given a large combinatorial context
and asked questions about that context:
\begin{itemize}
    \item Given a permutation $\sigma : \bof{1, \ldots, 64} \rightarrow \bof{1, \ldots, 64}$, compute $\sigma^k\of{x}$ for $k$ up to $64$.
    \item Given a function $f : \bof{1, \ldots, 8}^2 \rightarrow \bof{1, \ldots, 8}$ and a sequence of $64$ assignments of the form $x := 3$ or $x := f(y, z)$, evaluate a particular variable.
    \item Given a function $f : \bof{0, 1}^6 \rightarrow \bof{-1, 0, 1}$, answer questions of the form ``What is the sum of $f(x)$ over all $x$ matching the wildcard expression $0 \ast \ast \; 1 \ast \ast$?''
    \item Given a directed graph with $64$ vertices and $128$
    edges, find the distance from $s$ to $t$.
    \item Given a rooted forest on $64$ vertices,
    find the root of the tree containing a vertex $x$.
\end{itemize}

More detailed descriptions of the tasks
are available in \cref{app:tasks}.
We train each task using a curriculum of smaller
instances,
which is unrelated to our use of Iterated Amplification
(supervised learning also needs a curriculum
to learn these tasks in a reasonable amount of time
even given ground truth labels).

Rather than having a human perform the decomposition,
we provide a hard-coded algorithm $H$ which
decomposes each task
(though we minimize the number of times we call this algorithm).
Using these decompositions directly as a recursive
algorithm is not efficient for any of the tasks.

\subsection{Results}
\begin{figure}
\centering
\includegraphics[width=1.0\textwidth]{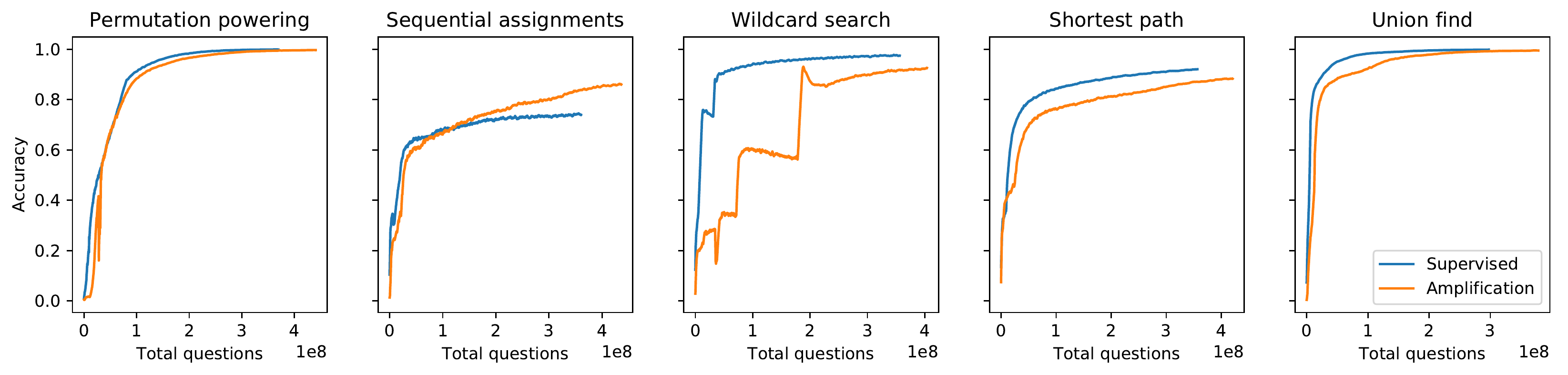}
\caption{Comparison between Iterated Amplification (orange)
and supervised learning from the ground truth data (blue).
To better illustrate the dynamics of training,
in \cref{app:dynamics} we compare the performance
of $X$ and $\amplifyp{X}$ separately for questions
requiring different recursion depths.}\label{fig:results}
\end{figure}

In order to evaluate Iterated Amplification,
we compare it to supervised learning from the ground truth data.
The results are presented in \cref{fig:results}.
Iterated Amplification is able to solve these tasks effectively
with at worst a modest slowdown,
achieving our main goal.

The purpose of amplification is to handle tasks
where an expert can perform decomposition but can't solve the task directly.
We don't expect amplification to solve those tasks as quickly as supervised learning.
Because we can learn these tasks \emph{almost} as quickly
as supervised learning from the ground truth,
we have achieved our main goal.

In addition to requiring modestly more training steps,
training with amplification requires about twice as much computation per question,
because we need to actually generate the training targets by running $\amplifyp{X}$.\footnote{Running $\amplifyp{X}$
requires calling both $X$ and $H'$ between 3 and 10 times.
We train on each $(Q, A)$ pair about 10 times before removing it from the dataset.
So the time required to generate a $(Q, A)$ pair is comparable to the total time spent training on it,
resulting in roughly twice the total computation per question.}
We trained in the same wall-clock time by generating data on a parallel worker.

Supervised learning required tens of millions of examples in order to learn these algorithms.
This would be a prohibitive cost if the examples were provided by an external expert.
In contrast, Iterated Amplification required tens of thousands of examples in order to learn the
much simpler decompositions (numbers for each task are in \cref{table:datause} in \cref{app:dynamics}).
The sample complexity will be much larger for realistic tasks,
where modeling $H$ might require (for example) a high-quality language model.
But the general phenomenon,
that learning decompositions can be much cheaper than learning complex behaviors directly,
is likely to hold for realistic tasks,
since performing decomposition is strictly easier
than solving the task without the aid of $X$.

\subsection{Simplifications}
\label{sec:future}

These experiments make several important simplifications:
\begin{itemize}
    \item In our experiments questions can be algorithmically
    decomposed into subquestions, and we replace the human with a hand-coded algorithm.
    These experiments don't shed any light on whether humans can decompose interesting real world tasks,
    nor on whether it would be feasible to learn messy real world decompositions.
    \item We work in combinatorial domains where it it is possible to construct an algorithmic training signal.
    This makes it possible for us to cheaply run experiments and easily evaluate our performance, but
in the long run we care about tasks where it is not possible to construct an algorithmic or even human training signal.
    \item In our experiments $X$ is trained by supervised learning from $\amplify{X}$. In many important applications
    we suspect that we would learn a reward function from $\amplify{X}$ and then train $X$ to maximize that reward function.
    \item In order for Iterated Amplification
to succeed,
the question distribution
$\dist$
needs to be broad enough to cover
not only the questions
we care about,
but also all of the subquestions
asked during the computation
of $\amplify{X}$.
The distribution also determines how the model
will allocate its capacity,
and so must be carefully chosen.
In our experiments,
we started with a distribution $\dist$ that could be used
directly.
In a more realistic setting, we might start with some distribution $\dist_0$ of questions
that have intrinsic interest,
and it would be the system designer's responsibility
to construct $\dist$ appropriately.
\end{itemize}
Removing these simplifications is a task for future work,
which will ultimately test the hypothesis that Iterated Amplification can be usefully applied to complex real-world tasks for which
no other training strategy is available.

\section{Discussion of decomposition in realistic domains}
\label{sec:decomposition}

Having successfully applied Iterated Amplification
to synthetic algorithmic problems,
the natural question is whether it can actually
be applied to complex real-world tasks that are ``beyond human scale.''
We leave a convincing demonstration to future work,
but we discuss here why we think this is likely.

The key assumption underlying Iterated Amplification
is that a human can coordinate multiple copies of $X$
to perform better than a single copy of $X$.

As an example, consider the problem of evaluating a proposed
design for a transit system.
Rather than forcing a single copy of $X$ to reach a snap judgment
about a proposed design,
we can have copies of $X$ evaluate many different considerations
(estimating costs, evaluating how well the system serves different populations, and so on).
A human can then decide how to aggregate those different considerations
(potentially with help from further copies of $X$).
We flesh out this example in more detail in \cref{app:decomposition}.

The problem of coordinating several copies of $X$
to outperform a single copy of $X$
is analogous to organizing a team of humans to outperform individual humans.
Fortunately,
there are several ways in which coordinating several copies of $X$
is easier than coordinating a team of humans:
\begin{itemize}
    \item We don't require that the collaboration be efficient,
    it just needs to help \emph{at all}.
    If ten agents working together perform ``10\% better''
    than a single agent on its own, then $\amplify{X}$
    still provides a useful training signal that we can use to improve $X$.
    \item The copies of $X$ don't need to run in parallel---each can start after the previous one has finished its task.
    Many tasks may be inherently difficult to parallelize,
    which is an obstacle for human collaboration but is
    fine for $\amplify{X}$.
    \item All of the copies of $X$ are trained exclusively to solve the problem they are given.
    We don't need to manage incentives,  politics, or conflicting preferences, which are common difficulties in human organizations.
\end{itemize}
Despite these difficulties,
human organizations are often able to significantly outperform individual humans in many domains,
supporting our key assumption.

\section{Conclusion}

We have shown that Iterated Amplification can successfully solve algorithmically complex tasks
where there is no external reward function and the objective is implicit in a learned decomposition.
This offers hope for applying ML in domains where we cannot compute a suitable objective,
even with human help---as long as humans are able to decompose a task into simpler pieces.
If we can realize this hope,
it will be an important step towards expanding
the reach of ML and
addressing concerns about the long-term impacts of AI
by reducing our reliance simple but inaccurate proxies
for complex implicit objectives.

%
%
%

\bibliography{references}

%

\newpage\appendix
\section{Training dynamics}
\label{app:dynamics}

\Cref{table:datause} shows how many queries
were collected over the course of training for each of our tasks.

\begin{table}
\centering
\caption{Total number of queries to the decomposition oracle $H$ for each task.
These experiments involve algorithmic tasks where these decompositions are straightforward;
the sample complexity will likely be much larger for more realistic tasks.}\label{table:datause}
\begin{tabular}{|c|ccccc|}
\hline
Task & \makecell{Permutation\\powering} & \makecell{Expression\\evaluation} & Union find & Shortest path & \makecell{Wildcard \\ search} \\
\hline
Oracle calls & 7,000 & 6,000 & 20,000 & 24,000 & 10,000
\\
\hline
\end{tabular}
\end{table}

\begin{figure}[h]
\centering
\includegraphics[width=0.9\textwidth]{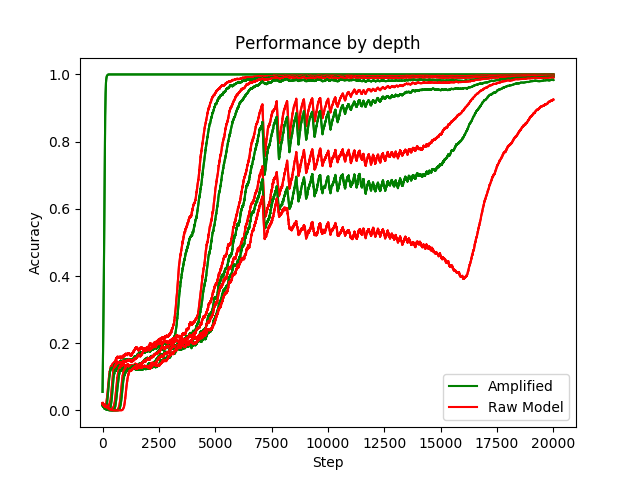}
\caption{
Performance at the permutation powering task,
for instances requiring different recursion depths.
The green curves are the accuracy of $\amplifyp{X}$;
the red curves are the accuracy of $X$,
The curves further to the right are higher depths,
i.e. larger powers of the permutation.
For example, the leftmost green curve
is the performance of $\amplifyp{X}$ at squaring or cubing
permutations,
which quickly converges to $1$ because
this task can be accomplished without $X$'s help. The rightmost red curve is the performance of $X$ on powers between $32$ and $63$.}\label{fig:dynamics}

\end{figure}

\Cref{fig:dynamics} illustrates
how $X$ ``chases'' the moving target $\amplifyp{X}$
over the course of training,
separating out performance for separate recursion depths $d$.

$X$ always has lower accuracy than $\amplifyp{X}$,
because it is being trained to imitate it.
$\amplifyp{X}$ for tasks at depth $d+1$
has slightly lower accuracy
than $X$ does for tasks at depth $d$,
because $\amplifyp{X}$ answering a question
correctly at depth $d+1$
requires $X$ to answer several questions
correctly at depth $d$.

The sawtooth pattern and decreases
in accuracy are an artifact of the curriculum.
In contrast with \cref{fig:results},
\cref{fig:dynamics}
shows the performance on the maximum difficulty of tasks yet encountered
by the model.
Each time the difficulty of the task is increased, the accuracy of the model drops.

\section{Example decomposition}
\label{app:decomposition}

Consider the task of comparing two designs for a public transit system.
We could train an AI to imitate human judgments, but human judgments may be very far from optimal.
Or we could try to collect enough information about the long-term health of transit systems to train an AI to predict long-term outcomes,
but we can only collect new datapoints with a ten-year delay.
This is the kind of task which we might want to solve with Iterated Amplification.

Many subquestions could help the human reach a conclusion about which design is better:
\begin{itemize}
    \item Compare the usefulness of the two designs.
    \item Compare the total cost of the two designs.
    \item Search for any risks or possible problems with each design so that they can be evaluated more carefully.
\end{itemize}
Combining the answers to these subquestions requires making value judgments, for example about how we should quantify and compare benefits,
what kinds of risks we are willing to accept,
and so on.
Our hope is for $X$ to learn about these value judgments
at the same time that it learns to make sophisticated decisions.
Eventually, $X$ can also help with the task of aggregating different considerations.

Each of these subquestion can itself be further divided,
facilitating the use of Iterated Amplification:
\begin{itemize}
    \item Comapre the usefulness of the two designs.
    \begin{itemize}
        \item Identify and evaluate particular needs that each design may serve well or poorly. \begin{itemize}
            \item Identify particular routes or groups of users who may be served well or poorly.
            \item For each, identify the most important considerations (predictability, reliability, routes served, cost) and assess how well each design meets those.
        \end{itemize}
        \item Forecast capacity of the system and likely usage.\begin{itemize}
            \item Evaluate current transit usage, correct for measurement issues, extrapolate trends.
            \item Evaluate capacity of the proposals across key routes and bottlenecks.
        \end{itemize} 
        \item Estimate general performance characteristics.\begin{itemize}
            \item Estimate how often the system will be unavailable and how common delays will be.
            \item Compare average speeds and waiting times.
            \item Compare plausible last-mile transit costs associated with each proposal.
        \end{itemize}
    \end{itemize}
    \item Compare the total cost of the two designs.
    \begin{itemize}
        \item Estimate the non-financial costs of the project \begin{itemize}
            \item Identify the most important effects on the landscape and space use within the city.
            \item Estimate the costs of disruption associated with construction and maintenance.
            \item Identify social consequences of the transit system.
        \end{itemize}
        \item Compare the likely construction costs of the two designs.\begin{itemize}
            \item Identify comparable projects and estimate their costs.
            \item Figure out how this project differs and how it's cost is likely to differ.
        \end{itemize}
        \item Compare the maintenance costs over time. \begin{itemize}
            \item Identify categories of maintenance cost and estimate each of them separately.
            \item Search for comparable projects.
        \end{itemize}
        \item Decide how to trade off immediate costs vs distant costs.\begin{itemize}
            \item Estimate interest rates on debt and the feasibility of borrowing to fund such a project.
            \item Estimate the rates of return on other uses of funds.
            \item Evaluate our intrinsic time preferences.
        \end{itemize}
    \end{itemize}
    \item $\cdots$
\end{itemize}
We emphasize that this need not be an \emph{efficient} decomposition
in order to be suitable for Iterated Amplification---answers
to the subquestion just need to help \emph{at all}
on the original task.
As long as that's true, 
we can use a final copy of $X$
to answer the question in light of these subanswers.
This will outperform a copy of $X$
who didn't get to see the subanswers,
and can hence provide a useful
training signal for $X$ to improve.

If we had an external ground truth measure of quality
then we wouldn't need a human to propose this kind of decomposition
and we could instead allow an AI to search for whatever predictor
worked best.
However, if we don't have access to an external ground truth,
we can use this kind of decomposition to define the task.

\section{Task descriptions}
\label{app:tasks}

An overview of our five tasks is given in \cref{table:tasks}.
A task describes a context, a family of questions,
a decomposition strategy, and a set of \emph{primitive questions}.
The primitive questions are the mechanism by which $\amplify{X}$
is able to learn about the context.
When $H$ asks a primitive question,
it immediately receives the correct answer
rather than being answered by $X$.

\begin{table}
\thisfloatpagestyle{empty}
\caption{The tasks on which we trained our algorithm.
All tasks are paremeterized by a difficulty $N$ that is increased from $8$ to $64$
over the course of training.
The decompositions marked with * are significant simplifications of
the full decompositions, which are described in \cref{sec:taskdetails}.}\label{table:tasks}
\begin{tabular}{l|c}
\hline
\multicolumn{2}{c}{\textbf{Permutation powering}} \\
\hline \\[1pt]
Context &  A permutation $\sigma$ of $N$ elements. \\[5pt]
Questions & What is $\sigma^k\of{x}$? (for $2 \leq k < 64$) \\[5pt]
Primitive questions & What is $\sigma\of{x}$? \\[5pt]
Decomposition & 
\makecell{
$\sigma^{2k}\of{x} = \sigma^k\of{\sigma^k\of{x}}$ \\
$\sigma^{2k+1}\of{x} = \sigma\of{\sigma^k\of{\sigma^k\of{x}}}$\\
} \\[10pt]
\\[1pt]
\hline
\multicolumn{2}{c}{\textbf{Sequential assignments}} \\
\hline \\[3pt]
Context &
\makecell{
A function $f : \bof{1, 2, \ldots, 8} \times \bof{1, 2, \ldots, 8} \rightarrow \bof{1, 2, \ldots 8}$ \\
A sequence of $N$ definitions of the form $x := f(y, z)$ or $x := 7$
} \\[10pt]
Questions & What is the value of $x$? \\[5pt]
Primitive questions &
\makecell{
What is the definition of $x$? \\
What is $f(a, b)$?
} \\[10pt]
Decomposition &
\makecell{
Look up definition of $x$. \\
If $x := f(y, z)$:
evaluate $y$, evaluate $z$, and look up $f(y, z)$.
} \\[10pt]
\\[1pt]
\hline
\multicolumn{2}{c}{\textbf{Union find}} \\
\hline \\[3pt]
Context & 
\makecell{
An undirected forest with $N$ vertices, $\sqrt{N}$ connected components,\\
and one vertex assigned a label in $\bof{1, \ldots, 8}$ in each component.
} \\[10pt]
Questions &
\makecell{
What is the unique label in the component containing $x$? \\
What is a vertex on the path from $x$ to a labeled vertex? \\
How far is $x$ from a labeled vertex?
} \\[15pt]
Primitive questions &
\makecell{
What is a random neighbor of $x$? \\
Are $x$ and $y$ connected? \\
Is $x$ assigned a label, and if so what is it?
} \\[15pt]
Decomposition\textsuperscript{*} &
\makecell {
What is a vertex on the path from $x$ to a labeled vertex? \\
If $y$ is the result: what is the unique label in the component containing $y$?
} \\[10pt]
\\[1pt]
\hline
\multicolumn{2}{c}{\textbf{Wildcard search}} \\
\hline \\[3pt]
Context & A function $f : \bof{0, 1}^6 \rightarrow \bof{-1, 0, 1}$ with $N$ non-zero values. \\[5pt]
Questions & What is $\sum f\of{x}$ over $x$ matching a wildcard expression (e.g. $0 \ast \ast \; 1 \ast 0$)? \\[5pt]
Primitive questions & What is $f(x)$? \\[5pt]
Decomposition & \makecell{Fill in $0$  for the first $*$ in the wildcard. \\
Fill in $1$ for the first $*$ in the wildcard. \\
Add the results}\\[15pt]
\\[1pt]
\hline
\multicolumn{2}{c}{\textbf{Shortest path}} \\
\hline \\[3pt]
Context & A directed graph with $2N$ edges and $N$ vertices. \\[5pt]
Questions &
\makecell{
What is the distance from $x$ to $y$? \\
What is the first vertex on the path from $x$ to $y$?
} \\[10pt]
Primitive questions &
\makecell{
What is a random neighbor of $x$? \\
Is there an edge from $x$ to $y$?
} \\[10pt]
Decomposition\textsuperscript{*} &
\makecell{
What is the first vertex on the path from $x$ to $y$? \\
If $z$ is the result: what is the distance from $z$ to $y$?
}
\\[10pt]
\end{tabular}
\end{table}

\subsection{Representations}

All tasks other than wildcard search involve a domain with size $N$.
For each of these tasks, we introduce $8$ special identifiers $\bof{a, b, c, d, e, f, g, h}$,
and represent elements from the domain as a pair of identifiers.

For sequential assignments, union finding, wildcard search, and shortest path,
we represent each integer with a separate symbol.
In the case of wildcard search this results in a very large vocabulary,
most of which is used extremely rarely and simply ignored by the model.
For permutation powering, we represent the exponent $k$ in binary.

In each domain, we can unambiguously represent facts
as a sequence of elements from the domain.
We represent function values $f(x)=y$ as the pair $xy$,
$x=f(y,z)$ as the triple $xyz$,
edges $(x, y)$ as the pair $xy$,
and so on.
Recall that elements from the domain are themselves represented as pairs,
which we simply concatenate.
For wildcard search, we simply omit zero values.

For union find and shortest path,
we preface each question with a unique symbol to disambiguate it.

We extend the vocabulary of each task with a special symbol ``?''
that is returned by $H$ whenever the recursive calls produce inconsistent or inconclusive results.

\subsection{Curriculum}

Each task has a size parameter $N$ that ranges from 8 to 64.
We begin training with the difficulty $8$.
Whenever $X$ achieves an accuracy of at least $85\%$
at predicting the $(Q, A)$ pairs in its dataset,\footnote{We exclude answers of ``?'',
since these are easy to correctly predict,
but don't indicate that the algorithm has mastered the task.}
we increment the difficulty.

At each point in training, we sample the task size $N$
to be equal to the difficulty with probability $0.5$,
and to otherwise be sampled from a power law between $8$ and $N$.

\Cref{fig:results} shows the performance on a held out test set,
which has $1/3$ of its examples at size $64$,
$1/3$ at size $8$,
and $1/3$ distributed uniformly between the two.
This distribution is chosen to give a useful representation
of progress throughout the learning process.
The graphs look essentially the same
(and would reach similar maximum accuracies) if we evaluate performance
on the most difficult tasks encountered so far,
except that it would no longer be meaningful to directly compare
different training runs that are at different difficulties
(and the curriculum introduces a sawtooth artifact as in \cref{fig:dynamics}).

\subsection{Detailed decompositions}
\label{sec:taskdetails}

The decompositions for union find and shortest path are somewhat more complex than the others.
We provide the full decomposition for shortest path here.
The decomposition for union find involves similar ideas.

\begin{itemize}
    \item What is the distance from $x$ to $y$?
    \begin{itemize}
        \item Test if $y$ is adjacent to $x$. If so, return $1$.\footnote{The distance from $x$ to $x$ is taken to be the length of the shortest cycle that contains $x$, rather than $0$.}
        \item $z \leftarrow$ What is the first vertex on the path from $x$ to $y$?
        \item Test if $z$ is adjacent to $x$. If not, return ?.
        \item $d \leftarrow$ What is the distance from $z$ to $y$?
        \item Return $d+1$.
    \end{itemize}
    \item What is the first vertex on the path from $x$ to $y$?
    \begin{itemize}
        \item $z \leftarrow$ What is the first vertex on the path from $x$ to $y$?
        \item Choose one at random:
        \begin{itemize}
            \item $w \leftarrow$ What is the first vertex on the path from $x$ to $y$?
            \item $w \leftarrow$ What is a random neighbor of $x$?
        \end{itemize}
        \item Test whether each of $z$ and $w$ are vertices adjacent to $x$.
        \item If neither of them is adjacent to $x$, return a random neighbor of $x$.
        \item If exactly one of them is adjacent to $x$, return that one.
        \item If both are adjacent to $x$, ask how far each of them is from $z$, and then return the one that is closer.
    \end{itemize}
\end{itemize}

\subsection{Task distributions}

For the permutation powering, union find, shortest path,
and wildcard search tasks,
the context is chosen uniformly at random from valid contexts.
For sequential assignments,
we sort the variables randomly,
assign each of the first $\sqrt{N}$ variables to one of $\bof{1, \ldots, 8}$
at random,
and let each subsequent variable be $f\of{y, z}$ for a random pair
$y, z$ of preceding variables.

For sequential assignments, shortest path, and union find,
we choose questions at random.
For wildcard search,
we randomly select the number of wildcards from $\bof{1, 2, \ldots, 6}$,
then we randomly generate a query with that many wildcards.
For permutation powering, we randomly choose one bit from $\bof{2, \ldots, 6}$
to be the leading bit of $k$.
We set that bit to be $1$ and set the other bits uniformly at random.
We selected these distributions $\mathcal{D}$
to ensure that every subquestion of a question draw from $\mathcal{D}$
is also given reasonable probability under $\mathcal{D}$.

\section{Model details and hyperparameters}
\label{app:details}

When computing $\amplifyp{X}$,
we use a Polyak averaging over a time horizon of 1000 batches,
rather than directly applying the current version of $X$.
This is analogous to the use of a target network in $Q$-learning.
Early experiments suggested Polyak averaging improved the stability of training,
and it does not materially slow down learning.

Our model closely follows the Transformer architecture \citep{transformer},
optimized with Adam.
We describe the model for completeness, along with our choice of hyperparameters.

\renewcommand{\ne}{d_{\mathrm{embed}}}
\newcommand{\nh}{d_{\mathrm{model}}}
All of our inputs are sets of $8$ to $128$
sentences,
each of which is a sequence
of $2$ to $8$ tokens
from a vocabulary of size $10$ to $30$.
For example,
we represent a graph as a list of pairs of vertices,
and we represent each vertex as a pair of tokens
from a fixed vocabulary of size $8$.

We process each sequence by embedding each token
in a space of dimension $\ne = 100$,
concatenating the embeddings,
and then applying a learned linear transformation
into a space of dimension $\nh = 512$.
Note that there is no position encoding,
because we process unordered sets of facts.

We then process a set of sentences
by applying a sequence of $N$ identical layers.
Each layer implements the transformation $x \rightarrow z$:
\begin{align*}
y \leftarrow \layernorm{x + \attention{x}}\\
z \leftarrow \layernorm{y + \batchnorm{\mlp{y}}}
\end{align*}
where $\attention{}$ is multi-headed attention with $8$
heads of size $\nh/8$, $\mlp{}$ is a two-layer
perceptron with hidden unit size $4\nh$ and ReLU activation,
$\layernorm{}$ is defined as usual,
and $\batchnorm{}$ normalizes across both the batch
and the different sentences.

We embed contexts using $N=6$ layers with self-attention.
Once we have processed a context,
we answer a batch of questions about that context
by using $N=3$ layers which attend
over the context embedding.
This is almost identical to the Transformer encoder/decoder,
except that a Transformer decoder
would also use self-attention (which is not appropriate
here since different questions are unrelated to one another).

This architecture outputs a single $\nh$ dimensional vector
for each answer.
To decode this into a sequence of tokens,
we train an autoregressive model that takes as input
the answer vector
and the sequence of tokens produced so far, and outputs the next token.
The output is allowed to either specify a token
directly or to specify an input token to copy,
as in pointer networks \citep{pointer}.
If the model chooses to copy an input symbol,
it outputs an attention mask to select a sentence
and a set of logits that are used to select which index
it wants to copy from that sentence.

Where possible we directly copied architecture choices from \citep{transformer}
because our goal was to focus on the training process
rather than architectural innovation.
We added batchnorm because it
significantly improved performance
during preliminary supervised experiments.

Each of our training runs involve between
$100,000$ and $200,000$ batches.
Each batch contains $50$ contexts.
The number of facts describing a context
varies from task to task,
and varies over the course of training as the task difficulty increases.
The number of questions per context
was the same as the number of facts.
By the end of training, this quantity was either $64$ or $128$
depending on the task.

The model was optimized with Adam,
with learning rate $10^{-5}$,
$\beta_2 = 0.98$, and gradient clipping.
These parameters were chosen based on early supervised
experiments.

\end{document}